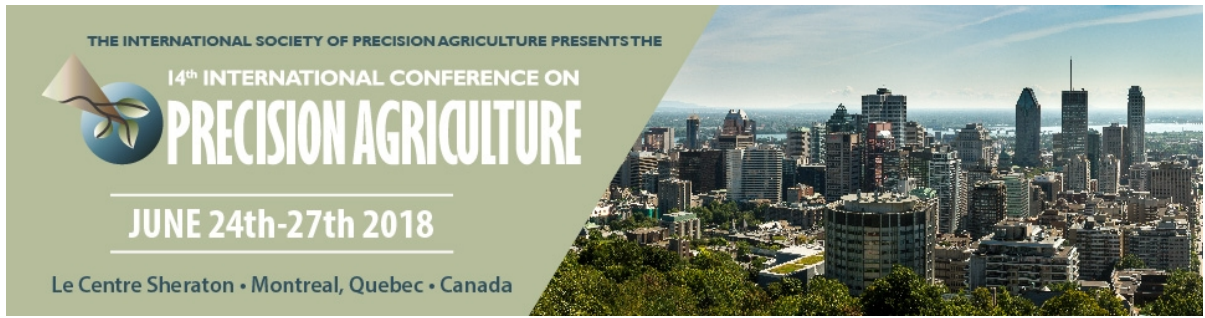

# Ground vehicle mapping of fields using LiDAR to enable prediction of crop biomass

**Martin Peter Christiansen[1], Morten Stigaard Laursen[1], Rasmus Nyholm Jørgensen[1], Søren Skovsen[1], René Gislum[2]**

[1]Department of Engineering, Aarhus University, Finlandsgade 22, 8200 Aarhus N, Denmark
[2]Department of Agroecology, Aarhus University, Forsøgsvej 1, 4200 Slagelse, Denmark

*Abstract* *Mapping field environments into point clouds using a 3D LIDAR has the ability to become a new approach for online estimation of crop biomass in the field. The estimation of crop biomass in agriculture is expected to be closely correlated to canopy heights. The work presented in this paper contributes to the mapping and textual analysis of agricultural fields. Crop and environmental state information can be used to tailor treatments to the specific site. This paper presents the current results with our ground vehicle LiDAR mapping systems for broad acre crop fields. The proposed vehicle system and method facilitates LiDAR recordings in an experimental winter wheat field. LiDAR data are combined with data from Global Navigation Satellite System (GNSS) and Inertial Measurement Unit (IMU) sensors to conduct environment mapping for point clouds. The sensory data from the vehicle are recorded, mapped, and analysed using the functionalities of the Robot Operating System (ROS) and the Point Cloud Library (PCL). In this experiment winter wheat (Triticum aestivum L.) in field plots, was mapped using 3D point clouds with a point density on the centimeter level. The purpose of the experiment was to create 3D LiDAR point-clouds of the field plots enabling canopy volume and textural analysis to discriminate different crop treatments. Estimated crop volumes ranging from 3500-6200 ($m^3$) per hectare are correlated to the manually collected samples of cut biomass extracted from the experimental field.*

*Keywords* *mobile robotics, canopy estimation, crop monitoring, point cloud, winter wheat mapping*

M. P. Christiansen
e-mail: mpc@eng.au.dk; martin.p.christiansen@gmail.com;

**Abbreviations**

| | |
|---|---|
| GNSS | Global Navigation Satellite System |
| LiDAR | Light Detection and Ranging |
| IMU | Inertial Measurement Unit |
| PCL | Point Cloud Library |
| RGB | Red, Green and Blue |
| ROS | Robot Operating System |
| RTK | Real Time Kinematic |
| UTM | Universal Transverse Mercator |
| WGS | World Geodetic System |

## Introduction

There is a current political demand to develop and implement a novel nitrogen application strategy in conventional Danish agriculture. The purpose of the new strategy is to optimize utilization of applied nitrogen to avoid negative environmental impact (Jespersen 2013). One solution could be to apply nitrogen according to current and predicted final nitrogen uptake of the crop. Measurement of nitrogen uptake (kg ha$^{-1}$) involves estimation of crop biomass and nitrogen status (e.g. percent nitrogen in crop biomass). Based on this estimation we can predict the necessary amount of nitrogen to apply in order to reach the warranted final nitrogen uptake at harvest. The future scenario is to estimate nitrogen uptake (kg nitrogen ha$^{-1}$) at predetermined growth stages and apply the amount of nitrogen needed to reach a predefined total nitrogen uptake (kg nitrogen ha$^{-1}$) at harvest.

In recent years both ground vehicle such as tractors (Rosell et al. 2009), (Palleja et al. 2010), (Reiser et al. 2018), and unmanned aerial vehicle (UAV) (Saari et al. 2011), (Sankey et al. 2017) systems have been used to collect sensory data from broad-acre fields automatically. Both ground vehicle and UAV solutions should be considered when developing new systems and applications for precision agriculture. A method to produce 3D representations of the surroundings is Light Detection and Ranging (LiDAR). LiDAR uses differences in laser light return time to measure representations of the surroundings.

In agriculture, LiDAR and hyperspectral sensory data have been combined to monitor vegetation (Sankey et al. 2017). The UAV with LiDAR and hyperspectral sensors was flying at an altitude of 70 m and a point density of 50 points/m$^2$. A lower flight altitude would provide an increase in spatial resolution of the surroundings, but draft from UAV propellers, can affect the monitored crop and thereby impact the result. The authors of this paper have previously evaluated a UAV-LiDAR based solution for mapping crop fields (Christiansen et al. 2017), where a DJI Matrice-100 was flown at an altitude of 3 meters. The main challenges with the UAV-LIDAR solution were the battery consumption when flying at low altitudes, if large areas need to be covered.

The UAV-LIDAR considerations makes it relevant to experiment with a ground-based vehicle solution to cover the field if LiDAR are needed for mapping crop fields. LiDAR sensors have been used in agriculture on ground-based vehicles for mapping soil (Jensen et al. 2014) and crops (Andújar et al. 2013), (Reiser et al. 2016). LiDAR mapping data can be used to evaluate the impact of agricultural production methods (Jensen et al. 2017). Orchards also tend to be an area of research where LiDAR have been used for mapping and monitoring (Jaeger-Hansen, Griepentrog, and Andersen 2012), (Underwood et al. 2015), (Andújar et al. 2016). Based on previous experience and the presented literature research we have chosen to evaluate the ground-based vehicle platform as a solution.

It was chosen to build upon the before mentioned UAV-LiDAR system and mount it on a vehicle as illustrated in Fig. 1, to have a point of comparison. The purpose of the experiment was to create 3D LIDAR point-clouds of a winter wheat field enabling canopy volume and do a comparison to biomass samples collected afterwards. The intention is also to evaluate the current vehicle LiDAR setup ability to be implemented in a possible future fully automated system for applying nitrogen. This paper also looks at the results in the perspective of the emerging LiDAR solutions for the automotive industry and their potential applications in a next generation system.

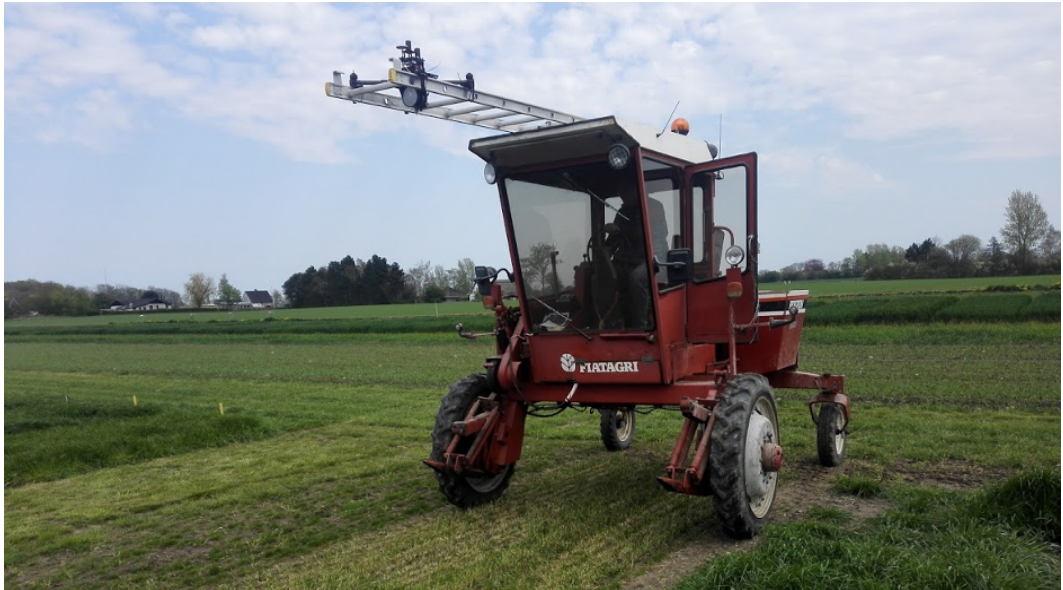

Fig. 1 Fiatagri Hesston 6500 used in the experiment to record the LiDAR data

## Materials

### Winter wheat experimental field

An experiment with winter wheat (Triticum aestivum L.) cv. Benchmark was established at Aarhus University, Flakkebjerg (WGS84: Lat 55.32729°, Lon 11.38846°), as illustrated in Fig. 2. The experimental field was seeded in autumn 2016. The planned plot size (crop parcel) was 2.28 × 8 m for an individual experiment. The experimental design was documented in (Christiansen et al. 2017), and involved four replicate crop parcels, for 21 Nitrogen application strategies randomized within each block. In total the experimental field consists of 84 crops parcels with an additional 19 to protect from wind and to separate experiments in the crop parcels. The separation between the crop parcels is made so the tires of the vehicle illustrated in Fig. 1, can driver over the crops and allow the LiDAR to scan the plants.

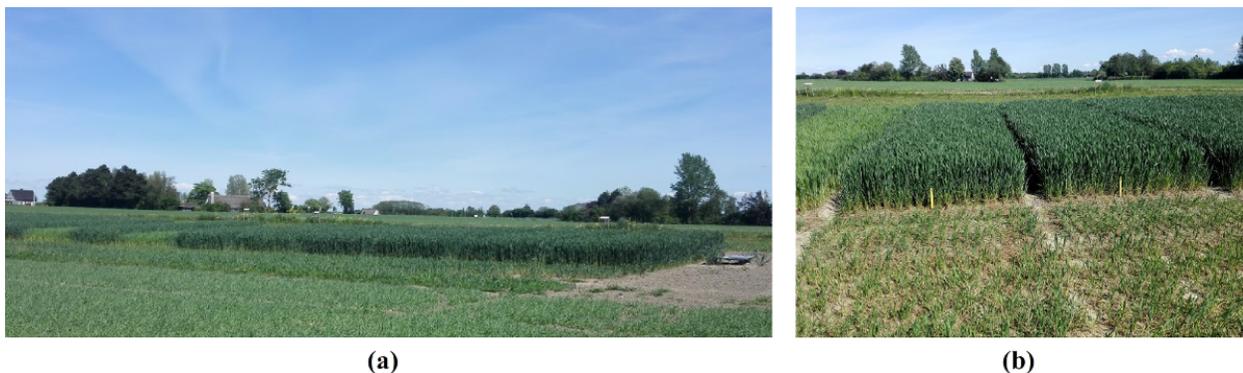

(a)  (b)

Fig. 2 The experimental field used for recording the sensor data. (a) The experimental crop parcels were sown in parallel rows facing perpendicular to the north. (b) Each crop parcel was seeded with 19 rows of winter wheat with 0.12 m distance between them. The separation between the crop parcels ensures the vehicle can drive over them.

## Vehicle and sensor setup

The vehicle setup is based on a HESSTON 6500 swather (Fiat Trattori, Turin, Italy) without the cutting implement, illustrated in Fig. 1. The sensor setup was specifically designed for mapping agricultural field areas, by allowing for simultaneous recording from multiple sensor sources. The sensor setup consists of a Velodyne VLP-16 LiDAR (San Jose, CA, USA), a color camera from Point Grey Chameleon3 3.2 MP (Richmond, BC, Canada) with a Sony imx265 sensor (Tokyo, Japan), a Trimble BD920 GNSS (Sunnyvale, CA, USA) real-time kinematics global navigation satellite system (RTK-GNSS) module and a Vectornav VN-200 (Dallas, TX, USA) inertial measurement unit (IMU) with additional GNSS input. Both IMU and RTK-GNSS uses a MAXTENA M1227HCT-A2-SMA antenna (Rockville, MD, USA) to receive GNSS signals. By ensuring that IMU, GNSS and LiDAR sensory data are timestamped with the high precision GPS time, they can be merged after recording have been performed.

Our sensory setup documented in the previous research article (Christiansen et al. 2017) has been mounted on the vehicle in an arm extending out in front of the vehicle, as illustrated in Fig. 3. The sensor setup estimates the vehicle's current pose, which is reused to calculate the projection of the individual LiDAR beams in the environment. For the experimental configuration the LiDAR has vertical resolution of 2°, a horizontal/azimuth resolution of approximately 0.2° and a typical range accuracy of ± 0.03 m (Velodyne LiDAR 2017).

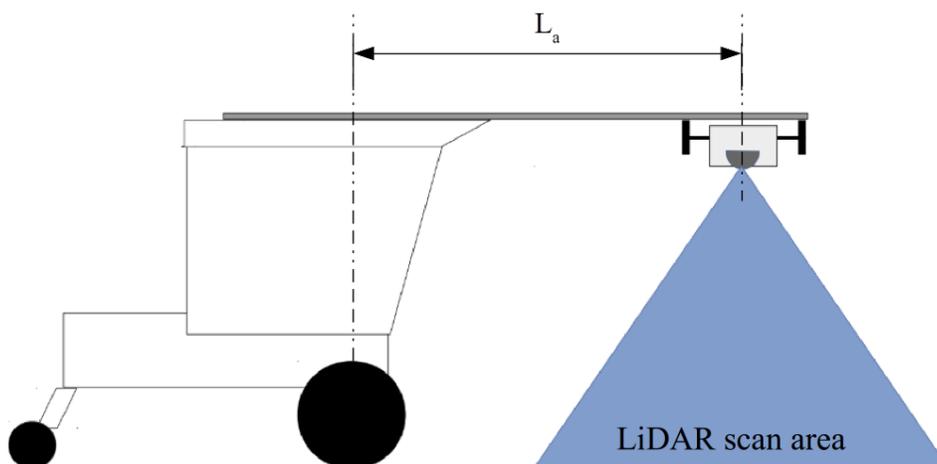

**Fig. 3 An illustration showing the placement of the UAV based sensor systems on the vehicle. The sensor system is mounted in arm out in front of the the vehicle, so the wheels and body does not cover the scan area of the LiDAR sensor. $L_a$ denotes the distance from the center frame of the sensor system to the center of the front wheels of the vehicle.**

It was chosen to orient the system in the same manner as when the UAV was flying. The LiDAR orientation differs significantly from the orientation seen on self-driving cars (Wolcott and Eustice 2014). The focus of the experiment is the crops and therefore a downward orientation of the LiDAR scan area was chosen. The intention with the selected orientation of the LiDAR was also to allow for comparison between the mapping results in the future.

An Odroid XU4 platform running Ubuntu 14.04 (Canonical Ltd, London, UK) armhf distribution and the robot operating system (ROS) (Quigley et al. 2009), (Carvalho et al. 2017) release Indigo, was used to record the sensor data. All data are stored in ROS files called bags that can record the published sensory information and allow for later playback. To increase the post-processing speed, the playback option was no utilized, but instead, the ROS bag was processed directly as data record file.

## Methods

### Recording date and procedure of the recording

The data logging with the sensor sytem and vehicle was performed on the 22 May 2017 in southwest Zealand, Denmark. The weather was sunny with wind speeds around 1 m/s.

The vehicle was manually driven along and over the crop parcels in the experimental field, as illustrated in Fig. 4.  The drive was instructed to cover both sides of the crop parcels, to ensure even coverage of the LiDAR scanning. The vehicle also drove over each fourth crop parcel in the experimental field.

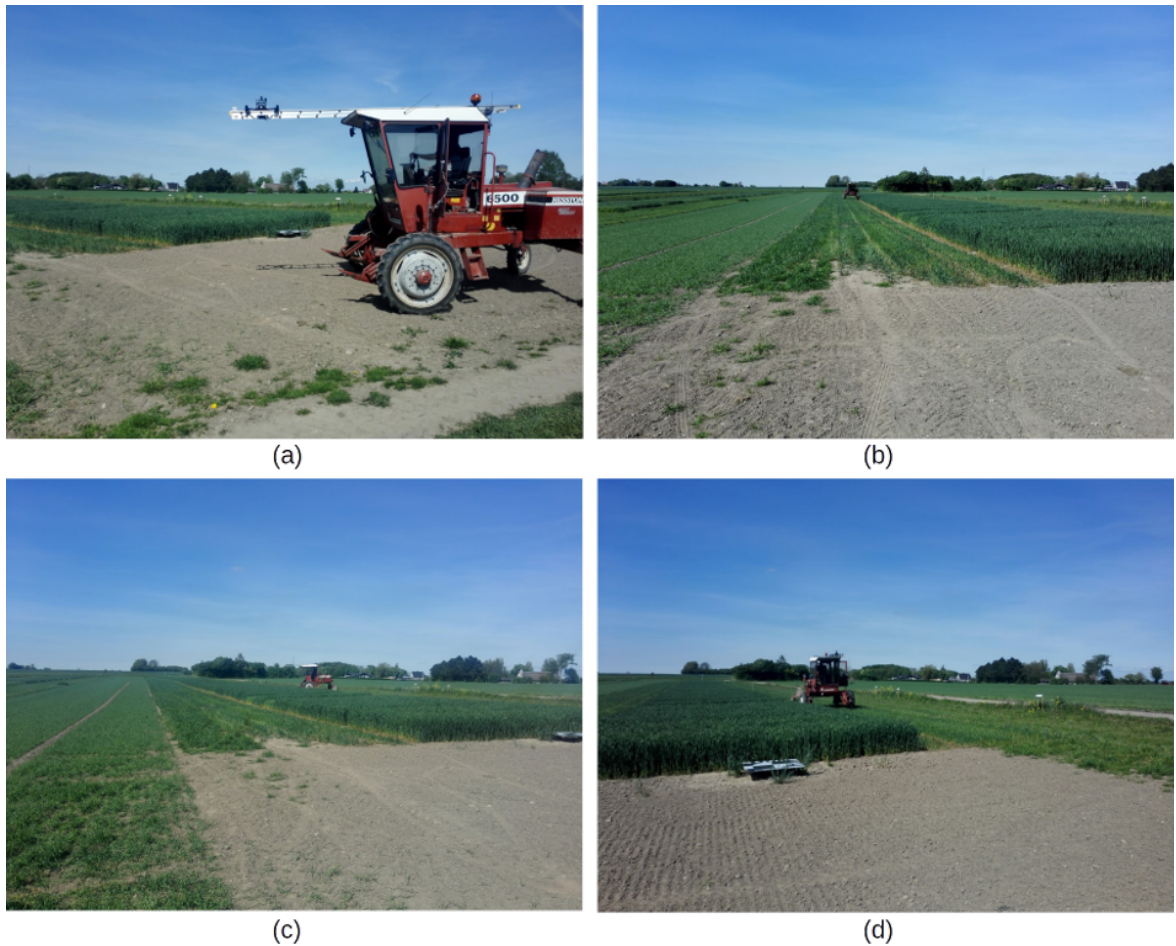

**Fig. 4 Images series documenting the recording procedure of the LiDAR data in the experimental field. (a) The starting position of the vehicle. (b/d) The recording is performed along both sides of the experimental field. (c) The vehicle drives over each fourth crop parcel.**

To provide reference location data, the corner point of each crop-parcel was logged using a Topcon GRS-1 GNSS-RTK receiver (Tokyo, Japan). The logged corner points were based on manual judgement of the beginning and end of the winter wheat parcels. An approximate gross parcel boundary was determined based on the logged crop-parcel corner points, and its position was used to extract individual parcel areas from the LiDAR data.

On the 23 May 2017, 0.5x0.5m areas was cut for all plants and extracted as biomass samples from 60 of the crop parcels. The position of the 60 parcels cut areas corners was also logged using the Topcon GRS-1 receiver. The goal with freshly cut plant samples, was to compare them against the estimated volume the LiDAR mapping produced. The assumption was that the winter wheat crop would not grow significantly overnight to impact the result of this comparison.

**Pose estimation in post-processing**

To use the LiDAR data for mapping crop parcels, the position and orientation (pose) of the sensor setup had to be estimated for each scan. The pose is estimated using GNSS and IMU data and utilize knowledge about the sensor systems placement on the vehicle. Each sensor on the UAV is represented by a reference frame relative to its position on the UAV (base-frame), as documented in (Christiansen et al. 2017). The base frame location of the system was chosen in the center of the RTK-GNSS antenna, as it allows direct mapping between the base-frame, the vehicle (vehicle frame) and the environment (global frame). The vehicle frame is placed above the front vehicle in the distance $L_a$ from the RTK-GNSS on the UAV. The GNSS position data is converted into universal transverse Mercator (UTM) coordinates to represent the global frame pose in Cartesian space. The VectorNav is used to determine the current roll and pitch angles, whereas a combination of IMU and RTK-GNSS is used to determine the yaw angle (vehicle heading). The yaw angle is calculated using an extended kalman filter that fuses information from IMU and RTK-GNSS based on the concept in (Bevly et al. 2009).

After processing the sensory data from IMU and GNSS, the pose of the base frame in the global frame for the recording period has been determined. The LiDAR scans are then mapped into the global reference frame (UTM32U) as a combined point-cloud.

**Processing of the collected LiDAR data**

To process the point-cloud data stored in ROS, we utilized the Point Cloud Library (PCL). Using PCL, all LiDAR scans were stored in the same point cloud based on their estimated relative position. Relative UTM pose coordinates were used for PCL processing because many PCL data containers utilize single-precision floating-point (Hsieh 2012).

An algorithm is used to distinguish between plant and soil in the created point clouds of the experimental plots, as a first processing stage, as described in (Christiansen et al. 2017). The estimated crop and their structure are significant factors that we intend to use as an alternative approach to determine the biomass of the crops. Plant volume is estimated based on a voxel grid for filtering down the LiDAR point clouds.

# Results

Fig. 5 illustrates a section of the mapped LiDAR data collected with the vehicle.

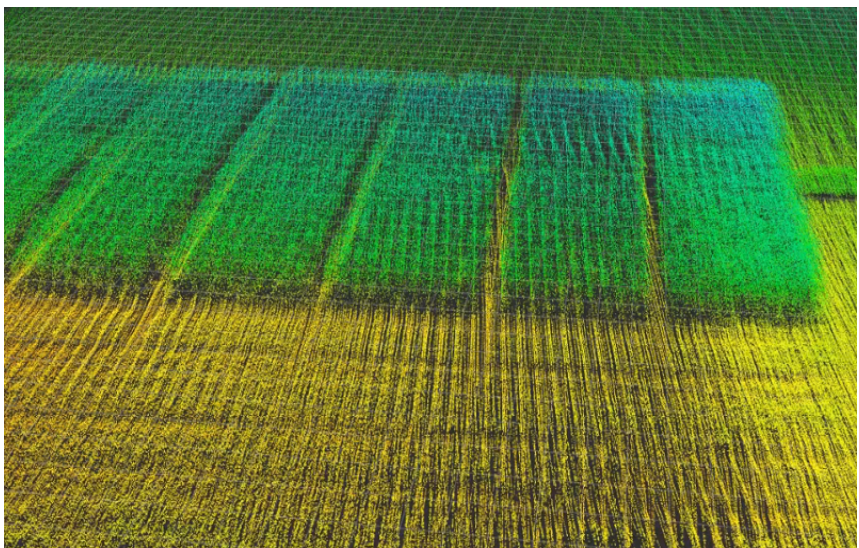

**Fig. 5** Subset illustrating the LiDAR point cloud created of the experimental field.

The comparison between biomass and estimated volume is illustrated in Fig. 6. A first and second order polynomial have been fitted to the data samples using least squares method.

The two polynomials was intended as models that could be used to predict future biomass level and are documented in equation (1) and (2).

$$b_{pw_1} = 8.850490\, e_v - 3205.553278 \tag{1}$$

$$b_{pw_2} = -0.00203\, e_v^2 + 28.47\, e_v - 49703.76 \tag{2}$$

where $b_{pw}$ represent the predicted biomass weight per hectare and $e_v$ the estimated winter wheat crop volume in an area. The two estimated polynomials have a coefficient of determination ($R^2$) of 0.551 and 0.574 respectively. These are rather low values for $R^2$ and likely stems from the distribution of values of the sample data.

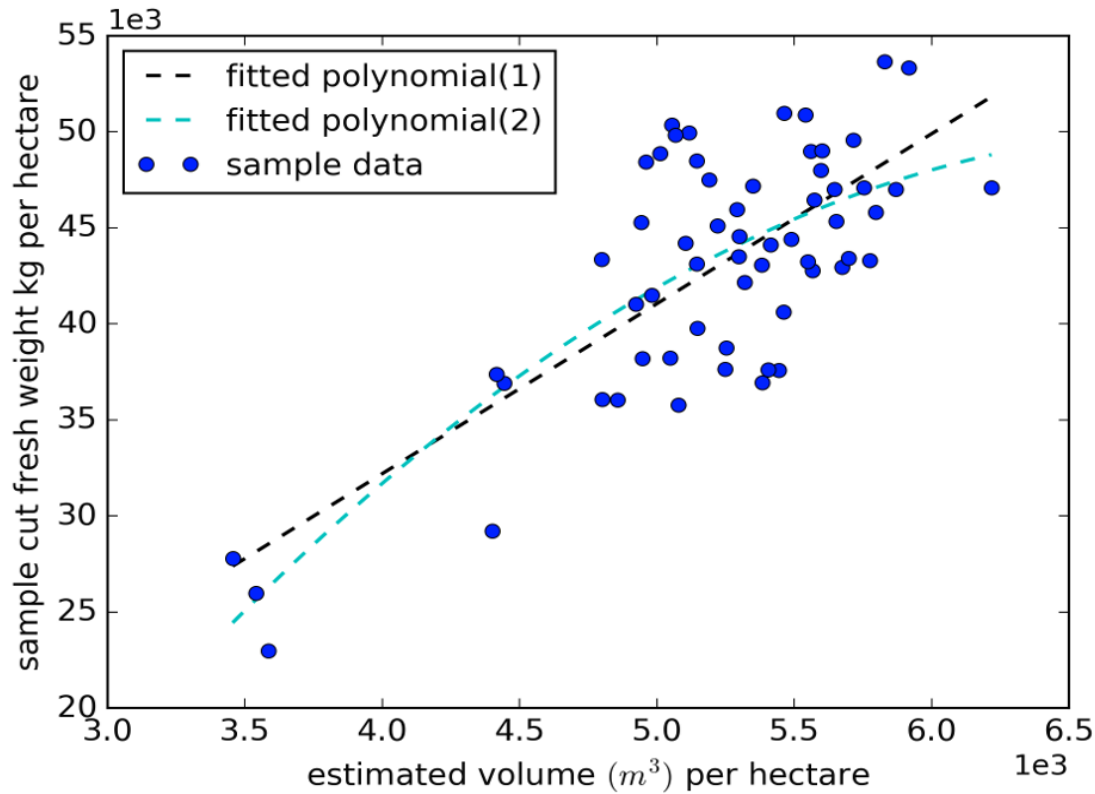

**Fig. 6 Comparison between fresh winter wheat biomass sample and estimated volume of the areas where the samples were collected.**

## Discussion

By estimating crop volume and comparing it to the biomass from extracted samples, we have shown that LiDAR point clouds could be used to predict biomass levels. Our results illustrate that the sensor system can be used to collect spatial data from crop-fields, which can be post-processed to derive canopy volume estimates that can be correlated to biomass samples from the individual crop parcels. Sample cut areas were extracted based on known GNSS reference corner points for the crop parcels, and crop volumes were calculated based on the LIDAR point cloud. The estimated volumes of the cut sample areas were in the range of 3500-6200 ($m^3$) per hectare. The estimated volumes and their matching fresh cut biomass samples have been correlated and model with a second order polynomial for allow for future prediction winter wheat biomass based on LiDAR data.

To produce more conclusive result additional samples need to be collected for different dates in the early growth cycle. Currently the number of samples with values of biomass below 30000 kg per hectare is only four. The authors have determined that additional biomass samples needs to be collected from the lower range, to ensure a more robust correlation and model can be developed.

For an online system based on a ground vehicle, one could enhance the ground level estimates with information about dimensions of the vehicle. Since agricultural ground vehicles tend to drive on the same tracks in the field, they can be used as a reference to distinguish between soil and plant point cloud measurements.

At the writing of this article, the price of a new VLP-16 LiDAR is around 4000 US dollars. In the future, solid state and flash LiDAR solutions (Chesworth and Huddleston 2018), , will become available for the consumer market, that could replace our current VLP-16 LiDAR. The upcoming generations of LiDAR would mean lower prices for any future iterations of the system or any commercial products providing the same service. Flash LiDAR acquires a single scan much like an ordinary camera (Thakur 2016) and thereby remove the recording time lag between individual points seen in a traditional LIDAR scan.

## Conclusion

We introduced a mapping method for crops in broad-acre fields and illustrated how mapped LiDAR data could estimate the current level of biomass. The system design and utilized software components are made available, to allow adaption in similar projects. In this project, we mapped winter wheat with a row distance of 0.12 m using a ground vehicle into 3D LiDAR point clouds. Cut biomass samples from the experimental field have been correlated to their estimate volumes in the 3D LiDAR point cloud, these range from 3500-6200 ($m^3$) per hectare.

The collected data indicates it should be possible to use reference LiDAR point cloud data volume estimates to biomass samples, to make predictions of future biomass level in a winter wheat field. In theory, a LiDAR-based system able to predict biomass level directly in a field could be used to online adjust the amount of fertilizer applied. Further referenced biomass samples from the growth cycle of winter wheat need to be collected before a fully automated system can be implemented.

### Acknowledgements

We would like to acknowledge Uffe Pilegård Larsen at Research Centre Flakkebjerg for helping in connection with the experiments. Thanks are due to Kjeld Jensen at the University of Southern Denmark for providing technical support on the Trimble GNSS platform. This research was financially supported by the Intelligente VIRKemidler til reduktion af reduktion af kvælstofudvaskningen (VIRKN) project, funded by the Danish Ministry of Environment and Foods Grønt Udviklings- og Demonstrationsprogram (GUDP). The work presented here is also partially supported by the FutureCropping Project funded by Innovation Fund Denmark.